\documentclass[runningheads]{llncs}
\usepackage{graphicx}
\usepackage{amsfonts}
\usepackage{amsmath,mathrsfs}
\usepackage{mathtools}
\usepackage{color}
\usepackage{graphicx,float,subfig,epstopdf}
\usepackage[capitalise]{cleveref}
\graphicspath{{figures/}}
\definecolor{mypink1}{rgb}{0.858, 0.188, 0.478}

\author{Sarah Parisot\thanks{This work was supported by the European Union's Seventh Framework Programme (FP/2007-2013) / ERC Grant Agreement no. 319456.} \and Sofia Ira Ktena \and Enzo Ferrante \and Matthew Lee \and Ricardo~Guerrerro Moreno \and Ben Glocker  \and Daniel Rueckert}
\authorrunning{Parisot et al.}   

\institute{Biomedical Image Analysis Group, Imperial College London, UK }

\title{Spectral Graph Convolutions for Population-based Disease Prediction}

\begin{document}
\maketitle

\begin{abstract}
Exploiting the wealth of imaging and non-imaging information for disease prediction tasks requires models capable of representing, at the same time, individual features as well as data associations between subjects from potentially large populations. Graphs provide a natural framework for such tasks, yet previous graph-based approaches focus on pairwise similarities without modelling the subjects' individual characteristics and features. On the other hand, relying solely on subject-specific imaging feature vectors fails to model the interaction and similarity between subjects, which can reduce performance. In this paper, we introduce the novel concept of Graph Convolutional Networks (GCN) for brain analysis in populations, combining imaging and non-imaging data. We represent populations as a sparse graph where its vertices are associated with image-based feature vectors and the edges encode phenotypic information. This structure was used to train a GCN model on partially labelled graphs, aiming to infer the classes of unlabelled nodes from the node features and pairwise associations between subjects. We demonstrate the potential of the method on the challenging ADNI and ABIDE databases, as a proof of concept of the benefit from integrating contextual information in classification tasks. This has a clear impact on the quality of the predictions, leading to 69.5\% accuracy for ABIDE (outperforming the current state of the art of 66.8\%) and 77\% for ADNI for prediction of MCI conversion, significantly  outperforming standard linear classifiers where only individual features are considered. \end{abstract}
\section{Introduction}

Recent years have seen an increasing volume of medical image data being collected and
stored. Large scale collaborative initiatives are acquiring and sharing hundreds of
terabytes of imaging, genetic and behavioural data. With this novel wealth of imaging and non-imaging data, there is a need for models capable of representing potentially large populations and exploiting all types of information.  
Graphs provide a natural way of representing populations and their similarities. In such setting, each subject acquisition is represented by a node and pairwise similarities are modelled via weighted edges connecting the nodes. Such models provide powerful tools for population analysis and integration of non-imaging data such as manifold learning \cite{brosch_manifold_2013,wolz_nonlinear_2012} or clustering algorithms \cite{parisot_probabilistic_2016}. Nonetheless, all the available information is encoded via pairwise similarities, without modelling the subjects' individual characteristics and features.
On the other hand, relying solely on imaging feature vectors, e.g. to train linear classifiers as in \cite{abraham2016deriving}, fails to model the interaction and similarity between subjects. This can make generalisation more difficult and reduce performance, in particular when the data is acquired using different imaging protocols. Convolutional Neural Networks (CNNs) have found numerous applications on 2D and 3D images, as powerful models that exploit features (e.g. image intensities) and neighbourhood information (e.g. the regular pixel grid) to yield hierarchies of features and solve problems like image segmentation~\cite{havaei_brain_2017} and classification. 
The task of subject classification in populations (e.g. for diagnosis) can be compared to image segmentation where each pixel is to be classified. In this context, an analogy can be made between an image pixel and its intensity, and a subject and its corresponding feature vectors, while the pairwise population graph equates to the pixel grid, describing the neighbourhood structure for convolutions. 
However, the application of CNNs on irregular graphs is not straightforward. This requires the definition of local neighbourhood structures and node orderings for convolution and pooling operations \cite{niepert2016learning}, which can be challenging for irregular graph structures.
Recently, graph CNNs were introduced \cite{defferrard2016convolutional}, exploiting the novel concept of signal processing on graphs~\cite{shuman2013emerging}, which uses computational harmonic analysis to analyse signals defined on irregular graph structures. These properties allow convolutions in the graph spatial domain to be dealt as multiplications in the graph spectral domain, extending CNNs to irregular graphs in a principled way. Such graph CNN formulation was successfully used in~\cite{kipf2016semi} to perform classification of large citation datasets.

\noindent \textbf{Contributions.} In this paper, we introduce the novel concept of Graph Convolutional Networks (GCN) for brain analysis in populations, combining imaging and non-imaging data. Our goal is to leverage the auxiliary information available with the imaging data to integrate similarities between subjects within a graph structure. We represent the population as a graph where each subject is associated with an imaging feature vector and corresponds to a graph 
vertex. The graph edge weights are derived from phenotypic data, and encode the pairwise similarity between subjects and the local neighbourhood system. This structure is used to train a GCN model on partially labelled graphs, aiming to infer the classes of unlabelled nodes from the node features and pairwise associations between subjects. We demonstrate the potential of the method on two databases, as a proof of concept of the advantages of integrating contextual information in classification tasks. First, we classify subjects from the Autism Brain Imaging Data Exchange (ABIDE) database as healthy or suffering from Autism Spectrum Disorders (ASD). 
The ABIDE dataset comprises highly heterogeneous functional MRI data acquired at multiple sites. We show how integrating acquisition information allows to outperform the current state of the art on the whole dataset \cite{abraham2016deriving} with a global accuracy of 69.5\%. Second, using the Alzheimer's Disease Neuroimaging Initiative (ADNI) database, we show how our model allows to seamlessly integrate longitudinal data and provides a significant increase in performance to 77\% accuracy for the challenging task of predicting the conversion from Mild Cognitive Impairment (MCI) to Alzheimer's Disease (AD). The code is publicly available at \url{https://github.com/parisots/population-gcn}.

\section{Methods}
\begin{figure}[t]
\centering
\includegraphics[width=0.95\linewidth]{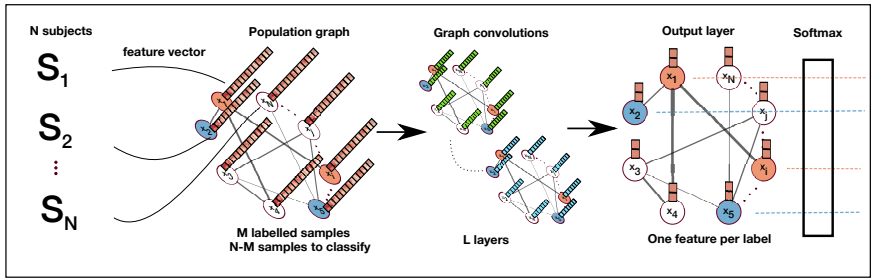}
\caption{Overview of the pipeline used for classification of population graphs using Graph Convolutional Networks.}
\label{fig:overview}
\end{figure}

We consider a database of $N$ acquisitions comprising imaging (e.g. resting-state fMRI or structural MRI) and non-imaging phenotypic data (e.g. age, gender, acquisition site, etc.). Our objective is to assign to each acquisition, corresponding to a subject and time point, a label $l \in \mathscr{L}$  describing the corresponding subject's disease state (e.g. control or diseased). To this end, we represent the population as a sparse graph $\mathcal{G}=\{\mathcal{V},\mathcal{E}, W\}$ where W is the adjacency matrix describing the graph's connectivity. Each acquisition $S_v$ is represented by a vertex $v \in \mathcal{V}$ and is associated with a $C$-dimensional feature vector $\mathbf{x}(v)$ extracted from the imaging data. The edges $\mathcal{E}$ of the graph represent the similarity between the subjects and incorporate the phenotypic information.
The graph labelling is done in a semi-supervised fashion, through the use of a GCN trained on a subset of labelled graph vertices. Intuitively, label information will be propagated over the graph under the assumption that nodes connected with high edge weights are more comparable.
An overview of the method is available in Fig.~\ref{fig:overview}.

\subsection{Databases and Preprocessing}
\label{subsec:dataset}

We apply our model on two large and challenging databases for binary classification tasks. With the \emph{ABIDE database}, we aim to separate healthy controls from ASD patients and exploit the acquisition information which can strongly affect the comparability of subjects. Our goal on the \emph{ADNI database} is to predict whether an MCI patient will convert to AD. Our objective is to demonstrate the importance of exploiting longitudinal information, which can be easily integrated into our graph structure, to increase performance.     

The \textbf{ABIDE database} \cite{di2014autism} aggregates data from different acquisition sites and openly shares functional MRI and phenotypic data of 1112 subjects\footnote{http://preprocessed-connectomes-project.org/abide/}. We select the same set of 871 subjects used in~\cite{abraham2016deriving}, comprising 403 individuals with ASD and 468 healthy controls acquired at 20 different sites. To ensure a fair comparison with the state of the art~\cite{abraham2016deriving}, we use the same preprocessing pipeline \cite{craddock2013}, which involves skull striping, slice timing correction, motion correction, global mean intensity normalisation, nuisance signal regression, band-pass filtering (0.01-0.1Hz) and registration of the functional MRI images to MNI152 standard anatomical space. The mean time series for a set of regions extracted from the Harvard Oxford (HO) atlas~\cite{desikan2006automated} were computed and normalised to zero mean and unit variance. 
The individual connectivity matrices $S_1, ..., S_N$ are estimated by computing the Fisher transformed Pearson's correlation coefficient between the representative rs-fMRI timeseries of each ROI in the HO atlas. 

The \textbf{ADNI database} is the result of efforts from several academic and private co-investigators \footnote{http://adni.loni.usc.edu}. To date, ADNI in its three studies (ADNI-1, -GO and -2) has recruited over 1700 adults, aged between 55 and 90 years, from over 50 sites from the U.S. and Canada. 
In this work, a subset of 540 early/late MCI subjects that contained longitudinal T1 MR images and their respective anatomical segmentations was used. In total, 1675 samples were available, with 289 subjects (843 samples) diagnosed as AD at any time during follow-up and labelled as converters. Longitudinal information ranged from 6 to 96 months, depending on each subject. Acquisitions after conversion to AD were not included.  
As of $1^{st}$ of July 2016 the ADNI repository contained 7128 longitudinal T1 MR images from 1723 subjects. ADNI-2 is an ongoing study and therefore data is still growing. Therefore, at the time of a large scale segmentation analysis (into 138 anatomical structures using MALP-EM \cite{ledig2015}) only a subset of 1674 subjects (5074 images) was processed, from which the subset used here was selected. 

\subsection{Population graph construction}
\label{subsec:graph_construction}

The proposed model requires two critical design choices: 1) the definition of the feature vector $\mathbf{x}(v)$ describing each sample, and 2) modelling the interactions between samples via the definition of the graph edges $\mathcal{E}$.
We keep the feature vectors simple so as to focus on evaluating the impact of integrating contextual information in the classification performance. For the ABIDE data-set, we follow the method adopted by  \cite{abraham2016deriving} and define a subject's feature vector as its vectorised functional connectivity matrix. Due to the high dimensionality of the connectivity matrix, a ridge classifier is employed to select the most discriminative features from the training set. For the ADNI dataset, we simply use the volumes of all 138 segmented brain structures.  

The definition of the graph's edges is critical in order to capture the underlying structure of the data and explain the similarities between the feature vectors. We construct our sparse graph aiming to incorporate phenotypic information in our model, providing additional information on how similar two samples' feature vectors and labels are expected to be. Considering a set of $H$ non-imaging measures $\mathbf{M} = \{M_h\}$ (e.g. subject's gender and age), the population graph's adjacency matrix $W$ is defined as follows: 
\begin{equation}
W(v,w)= Sim(S_v,S_w) \sum_{h=1}^H \rho(M_h(v),M_h(w)),
\end{equation}
\noindent where, $Sim(S_v,S_w)$ is a measure of similarity between subjects, increasing the weights between the most similar graph nodes; $\rho$ is a measure of distance between phenotypic measures. Considering categorical data such as gender or acquisition site, we define $\rho$ as the Kronecker delta function $\delta$. For quantitative measures such as the subject's age, we define $\rho$ as a unit-step function with respect to a threshold $\theta$: 
$\rho(M_h(v),M_h(w)) = \begin{cases} 1 &\mbox{if } \vert M_h(v)-M_h(w) \vert < \theta  \\ 
0 & \mbox{otherwise} \end{cases}$

The underlying idea behind this formulation is that non-imaging complementary data can provide key information explaining correlations between subjects' feature vectors. The objective is to leverage this information, so as to define an accurate neighbourhood system that optimises the performance of the subsequent graph convolutions.
For the ABIDE population graph, we use $H=2$ non-imaging measures, namely subject's \emph{gender} and \emph{acquisition site}. We define $Sim(S_v,S_w)$ as the correlation distance between the subjects' rs-fMRI connectivity networks after feature selection, as a separation between ASD and controls can be observed within certain sites. The main idea behind this graph structure is to leverage the site information, as we expect subjects to be more comparable within the same site due to the different acquisition protocols. 
The ADNI graph is built using the subject's \emph{gender} and \emph{age} information. These values are chosen because our feature vector comprises brain volumes, which can strongly be affected by age and gender. The most important aspect of this graph is the $Sim(S_v,S_w)$ function, designed to leverage the fact that longitudinal acquisitions from the same subject are present in the database. While linear classifiers treat each entry independently, here we define $Sim(S_v,S_w) = \lambda$ with $\lambda > 1$
if two samples correspond to the same subject, and $Sim(S_v,S_w) = 1$ otherwise, indicating the strong similarity between acquisitions of the same subject. 

\subsection{Graph Labelling using Graph Convolutional Neural Networks}
\label{subsec:filtering}

Discretised convolutions, those commonly used in computer vision, cannot be easily generalised to the graph setting, since these operators are only defined for regular grids, e.g. 2D or 3D images. Therefore, the definition of localised graph filters is critical for the generalisation of CNNs to irregular graphs. This can be achieved by formulating CNNs in terms of spectral graph theory, building on tools provided by graph signal processing (GSP)~\cite{shuman2013emerging}.

The concept of spectral graph convolutions exploits the fact that convolutions are multiplications in the Fourier domain. The graph Fourier transform is defined by analogy to the Euclidean domain from the eigenfunctions of the Laplace operator. The normalised graph Laplacian of a weighted graph $\mathcal{G} = \{ \mathcal{V}, \mathcal{E}, W\}$ is defined as $\mathcal{L} = I_N - D^{-1/2} W D^{-1/2}$ where $I_N$ and $D$ are respectively the identity and diagonal degree matrices. Its eigendecomposition, $\mathcal{L}=U \Lambda U^T$, gives a set of orthonormal eigenvectors $U \in \mathbb{R}^{N \times N}$ with associated real, non-negative eigenvalues $\Lambda \in \mathbb{R}^{N \times N}$. The eigenvectors associated with low frequencies/eigenvalues vary slowly across the graph, meaning that vertices connected by an edge of large weight have similar values in the corresponding locations of these eigenvectors. 

The graph Fourier Transform (GFT) of a spatial signal $\mathbf{x}$ is defined on the graph $\mathcal{G}$ as $\hat{\mathbf{x}} \doteq U^T\mathbf{x} \in \mathbb{R}^{N}$, while the inverse transform is given by $\mathbf{x} \doteq U \hat{\mathbf{x}}$. Using the above formulations, spectral convolutions of the signal $\mathbf{x}$ with a filter $g_{\theta}=diag(\theta)$ are defined as 
$g_{\theta} \ast \mathbf{x} = g_{\theta}(\mathcal{L})\mathbf{x} = g_{\theta}(U \Lambda U^T)\mathbf{x} = U g_{\theta}(\Lambda) U^T \mathbf{x}$,

\noindent where $\theta \in \mathbb{R}^{N}$ is a vector of Fourier coefficients. Following the work of Defferrard et al.~\cite{defferrard2016convolutional}, we restrict the class of considered filters to polynomial filters $g_{\theta}(\Lambda) = \sum_{k=0}^{K}\theta_k \Lambda^k$. This approach has two main advantages: 1) it yields filters that are strictly localised in space (a $K$-order polynomial filter is strictly $K$-localised) and 2) it significantly reduces the computational complexity of the convolution operator. Indeed, such filters can be well approximated by a truncated expansion in terms of Chebyshev polynomials which can be computed recursively. 
Similarly to what is proposed in~\cite{kipf2016semi}, we keep the structure of our GCN relatively simple. It consists of a series of convolutional layers, each followed by Rectified Linear Unit (ReLU) activation functions to increase non-linearity, and a convolutional output layer. The output layer is followed by a softmax activation function~\cite{kipf2016semi}, while cross-entropy is used to calculate the training loss over all labelled examples. Unlabelled nodes are then assigned the labels maximising the softmax output.

\section{Results}

We evaluate our method on both the ADNI and ABIDE databases using a 10-fold stratified cross validation strategy. The use of 10-folds facilitates the comparison with the ABIDE state of the art \cite{abraham2016deriving} where a similar strategy is adopted. To provide a fair evaluation for ADNI, we ensure that the longitudinal acquisitions of the same subject are in the same fold (i.e. either the testing or training fold). 
We train a fully convolutional GCN with $L$ hidden layers approximating the convolutions with $K=3$ order Chebyshev polynomials. GCN parameters were optimised for each database with a grid search on the full database. For ABIDE, we use: $L=1$, dropout rate: 0.3, l2 regularisation: $5.10^{-4}$, learning rate: 0.005, number of features $C=2000$. The parameters for ADNI are: $L=5$, dropout rate: 0.02, l2 regularisation: $1.10^{-5}$, learning rate: 0.01, graph construction variables $\lambda = 10$ and $\theta=2$. 
The ABIDE network is trained for 150 epochs. Due to the larger network size, we train the ADNI network longer, for 200 epochs. 

\begin{figure}[t]
\centering
\subfloat[ABIDE accuracy]{\includegraphics[width=0.25\textwidth]{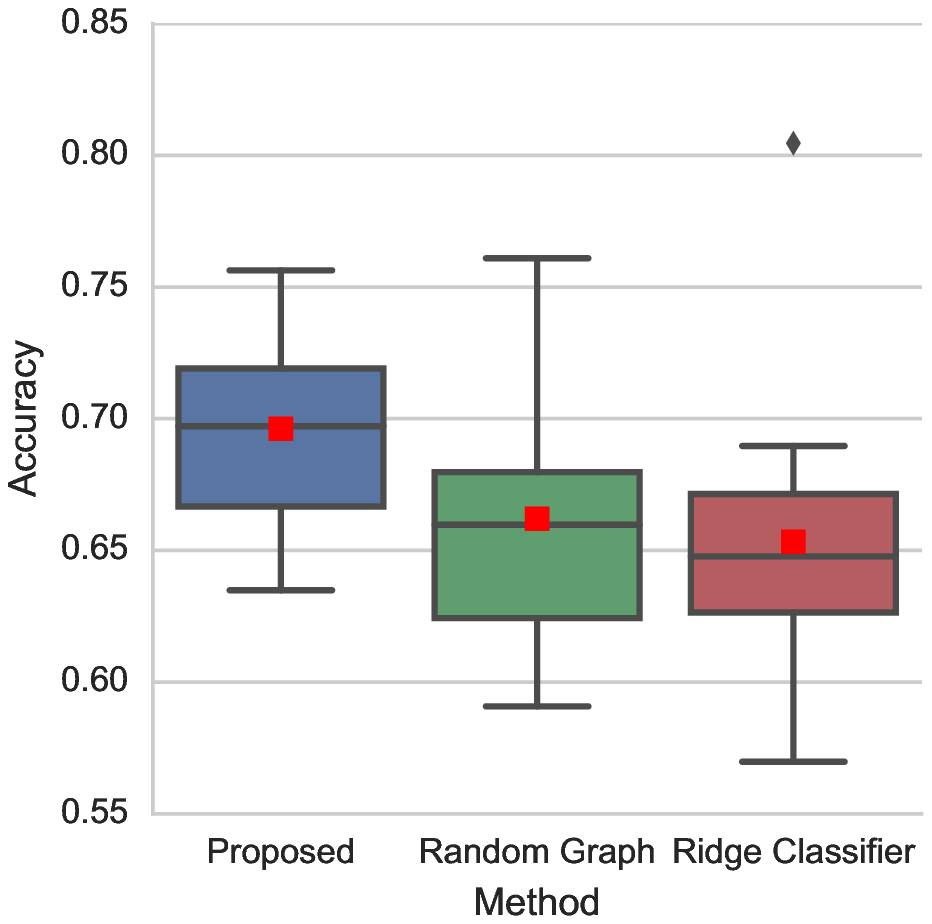}\label{subfig:ABIDEacc}}
\subfloat[ABIDE AUC]{\includegraphics[width=0.25\textwidth]{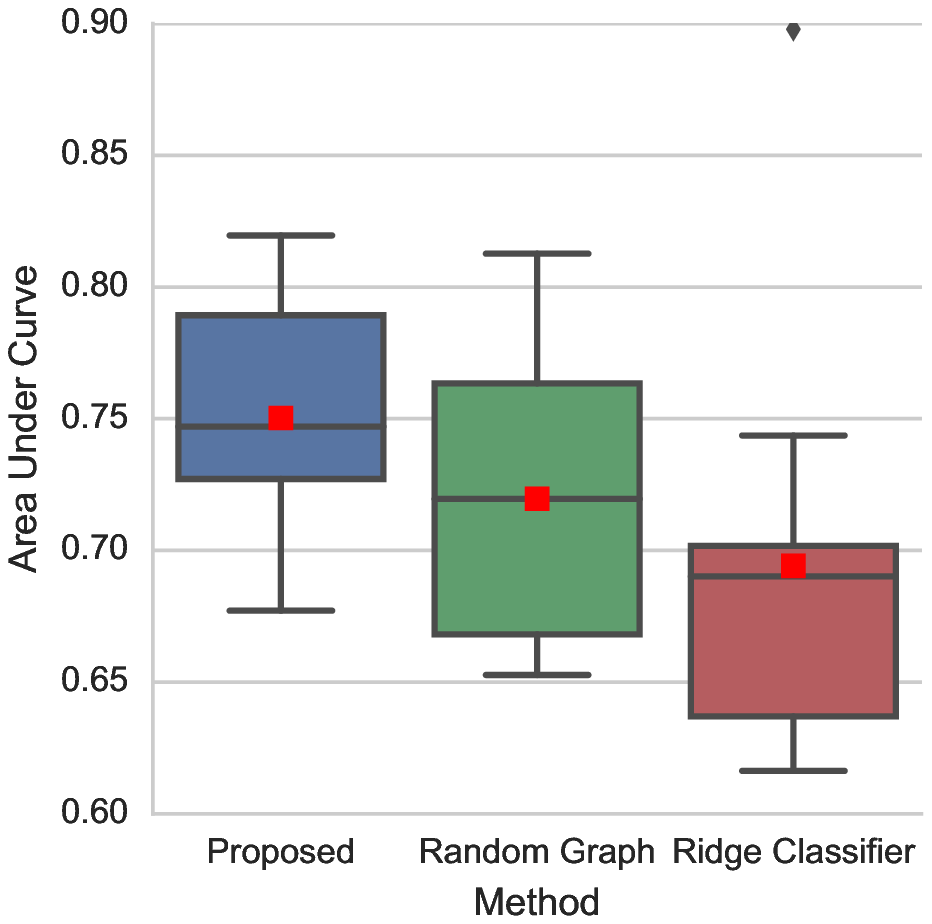}\label{subfig:ABIDEauc}}
\subfloat[ADNI accuracy]{\includegraphics[width=0.25\textwidth]{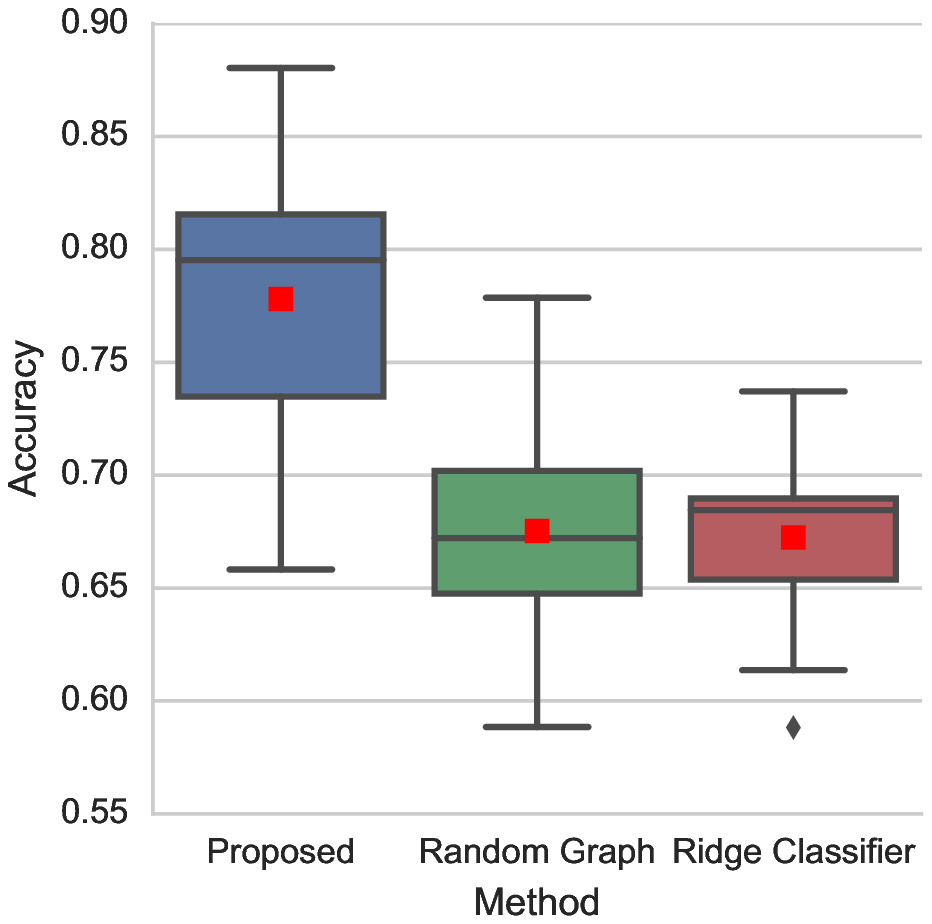}\label{subfig:ADNIacc}}
\subfloat[ADNI AUC]{\includegraphics[width=0.25\textwidth]{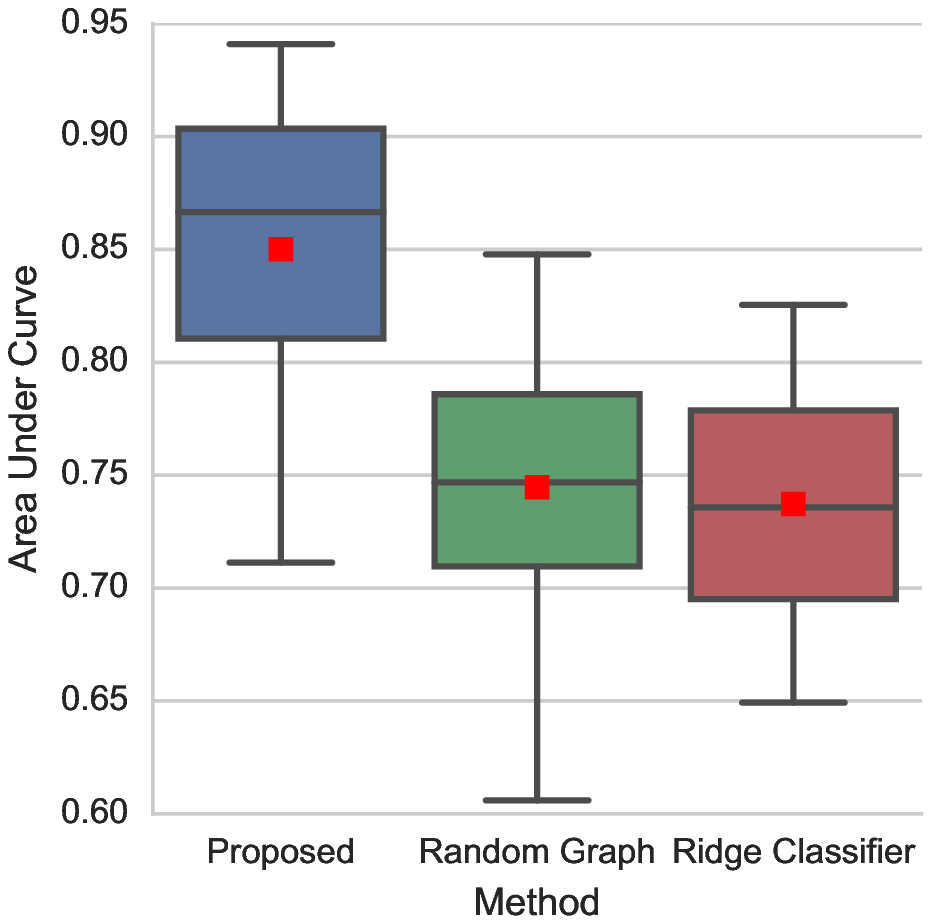}\label{subfig:ADNIauc}}
\caption{Comparative boxplots of the classification accuracy and area under curve (AUC) over all cross validation folds for the (a, b) ABIDE and (c, d) ADNI databases (MCI conversion task). The red dots correspond to the mean value.}
\label{fig:boxplots}
\end{figure}

We compare our results to linear classification using a ridge classifier (using the scikit-learn library implementation \cite{sklearn}) which showed the best performance amongst linear classifiers. We investigate the importance of the population graph structure by using the same GCN framework with a random graph support of same density. Comparative boxplots across all folds between the three approaches are shown in Fig. \ref{fig:boxplots} for both databases. GCN results (both with population and random graphs) are computed for ten different initialisation seeds and averaged. 
For both databases, we observe a significant ($p<0.05$) increase both in terms of accuracy and area under curve using our proposed method, with respect to the competing methods. The random support yields equivalent or worse results to the linear classifier. For ABIDE, we obtain an average accuracy of 69.5\%, outperforming the recent state of the art (66.8\%) \cite{abraham2016deriving}. Results obtained for the ADNI database show a large increase in performance with respect to the competing methods, with an average accuracy of 77\% on par with state of the art results \cite{tong2016}, corresponding to a 10\% increase over a standard linear classifier.

\section{Discussion}

In this paper, we introduced the novel concept of graph convolutions for population-based brain analysis. We proposed a strategy to construct a population graph combining image based patient-specific information with non-imaging based pairwise interactions, and use this structure to train a GCN for semi-supervised classification of populations. As a proof of concept, the method was tested on the challenging ABIDE and ADNI databases, respectively for ASD classification from a heterogeneous database and predicting MCI conversion from longitudinal information. Our experiments confirmed our initial hypothesis about the importance of contextual pairwise information for the classification process. In the proposed semi-supervised learning setting, conditioning the GCN on the adjacency matrix allows to learn representations even for the unlabelled nodes, thanks to the supervised loss gradient information that is distributed across the network. This has a clear impact on the quality of the predictions, leading to about 4.1\% improvement for ABIDE and 10\% for ADNI when comparing to a standard linear classifier (where only individual features are considered). 

Several extensions could be considered for this work. Devising an effective strategy to construct the population graph is essential and far from obvious. Our graph encompasses several types of information in the same edge. An interesting extension would be to use attributed graphs, where the edge between two nodes corresponds to a vector rather than a scalar. This would allow to exploit complementary information and weight the influence of some measures differently. Integrating time information with respect to the longitudinal data could also be considered. Our feature vectors are currently quite simple, as our main objective was to show the influence of the contextual information in the graph. We plan to evaluate our method using richer feature vectors, potentially via the use of autoencoders from MRI images and rs-fMRI connectivity networks.

\bibliographystyle{splncs03}
\bibliography{paper897}

\end{document}